# Advances in Exact Bayesian Structure Discovery in Bayesian Networks

**Mikko Koivisto**
HIIT Basic Research Unit, Department of Computer Science
Gustaf Hällströminkatu 2b, FIN-00014 University of Helsinki, Finland
mikko.koivisto@cs.helsinki.fi

## Abstract

We consider a Bayesian method for learning the Bayesian network structure from complete data. Recently, Koivisto and Sood (2004) presented an algorithm that for any single edge computes its marginal posterior probability in $O(n2^n)$ time, where $n$ is the number of attributes; the number of parents per attribute is bounded by a constant. In this paper we show that the posterior probabilities for *all* the $n(n-1)$ potential edges can be computed in $O(n2^n)$ total time. This result is achieved by a forward–backward technique and fast Möbius transform algorithms, which are of independent interest. The resulting speedup by a factor of about $n^2$ allows us to experimentally study the statistical power of learning moderate-size networks. We report results from a simulation study that covers data sets with 20 to 10,000 records over 5 to 25 discrete attributes.

## 1  INTRODUCTION

Learning the structure of a Bayesian network model is a central goal in many applications of Bayesian networks. In causal discovery, for example, one aims at the identification of (direct) causal relations among a set of attributes, which are qualitatively represented by edges of a Bayesian network (see, e.g., Cooper and Herskovits, 1992, Heckerman et al., 1999, Pearl, 2000, Spirtes et al., 2000).

Unfortunately, structure learning is generally hard, because (a) a finite number of observations may not reliably identify the true structure, and (b) evaluating the possible network structures—whose number generally grows superexponentially in the number of attributes—is computationally intractable. In practice, one often passes a relatively small data set to a heuristic evaluation algorithm. It is then difficult to quantify the uncertainty about the learning results: Although the uncertainty due to limited data and vague background knowledge can be rigorously expressed using standard statistical concepts (especially the Bayesian ones), the extra uncertainty induced by an algorithm with no performance guarantees is more complicated to gauge or control. Thus, it is desirable to extent the scope of exact algorithms, which completely avoid the extra layer of uncertainty.

This paper puts forward the Bayesian approach to structure learning (Buntine, 1991, Cooper and Herskovits, 1992, Friedman and Koller, 2003). We improve a recently found exact algorithm of Koivisto and Sood (2004), which computes the marginal posterior probability of any given subnetwork, e.g., a single edge, in $O(n2^n)$ time; here we assume that the indegree, i.e., the number of parents per attribute, is bounded by a constant. In Sections 2–4 we show how the probabilities for *all* potential edges can be simultaneously computed in $O(n2^n)$ total time. The resulting computational saving, by a factor of $n^2$, is very significant in practice (from two to three orders of magnitude); note that these algorithms are practical for networks of up to about $n = 25$ attributes. We achieve this improvement by using a technique that is analogous to the forward–backward method of hidden Markov models (see, e.g., Rabiner, 1989). Another ingredient is a fast Möbius transform algorithm, which we analyze carefully in Section 4, elaborating on some previous results (Kennes and Smets, 1991, Koivisto and Sood, 2004).

The improved exact algorithm allows us to study the statistical power of structure discovery in moderate-size networks. Given a model specification that fixes the number of attributes and the maximum number of parents per attribute, how many observations should one collect to reliably discover the network structure? Or, given a fixed data size, should one expect that a large proportion of the structure can be learned from the data? While it is obvious that the number of ob-servations plays a crucial role here, it is less clear how the statistical power behaves as a function of the maximum indegree and the number of states per attribute. In Section 5 we study the statistical power of edge discovery under varying settings (number of attributes, number of observations, maximum indegree, number of states per attribute).[1] Our study is complementary to a previous work by Husmeier (2003) that uses a Markov chain Monte Carlo method and focuses on very small data sets and large, yet very sparse, dynamic Bayesian networks.

We discuss some other potential applications of the presented exact algorithm in Section 6.

## 2 PRELIMINARIES

A Bayesian network (BN) over a vector of variables $x = (x_1, \ldots, x_n)$ specifies a probability distribution of $x$. The *network structure* of a BN encodes conditional independence assertions among the variables via a directed acyclic graph. We represent this graph as a vector $G = (G_1, \ldots, G_n)$ where each $G_i$ is a subset of the index set $V = \{1, \ldots, n\}$ and specifies the parents of $i$ in the graph. We use indexing with subsets: for example, if $S = \{i_1, \ldots, i_s\}$ with $i_1 < \cdots < i_s$, then $x_S$ denotes the vector $(x_{i_1}, \ldots, x_{i_s})$. Along the structure $G$, a BN factorizes the probability distribution of $x$ into a product of *local conditional distributions*. Usually these conditional distributions belong to some parametric family, and it is convenient to write the probability distribution of $x$ as

$$p(x|G, \theta) = \prod_{i=1}^{n} p(x_i | x_{G_i}, G, \theta),  \quad (1)$$

where $\theta$ contains the *parameters* of the local conditional distributions. This notation used here and henceforth supports the Bayesian approach that treats the network structure $G$ and the parameters $\theta$ as random variables (whenever their values are unknown).

BNs can be used to model multiple vectors $x[1], \ldots, x[m]$, called *data* and denoted by $\mathbf{x}$. In the Bayesian learning framework the vectors are judged to be exchangeable so that the probability distribution of the data, given the structure $G$, can be expressed as

$$p(\mathbf{x}|G) = \int \Big( \prod_{t=1}^{m} p(x[t]|G, \theta) \Big) p(\theta|G) \mathrm{d}\theta,$$

where $p(\theta|G)$ is a prior of the parameters, and each term $p(x[t]|G, \theta)$ factorizes as in (1).

To complete the Bayesian learning framework we introduce a prior of the network structure. We follow Friedman and Koller (2003) and augment the model with a new random variable, $\prec$, that specifies the linear order of the attributes.[2] Formally, $\prec$ is a linear order on the index set $V$, represented as a vector $(U_1, \ldots, U_n)$, where $U_i$ defines the predecessors of $i$ in the order, i.e., $U_i = \{j : j \prec i\}$; we may also write more completely $U_i^\prec$. If $S$ is a subset $V$ we let $\mathcal{L}(S)$ denote the set of linear orders on $S$. We say that a structure $(G_1, \ldots, G_n)$ is consistent with an order $(U_1, \ldots, U_n)$ if $G_i \subseteq U_i$ for all $i$; we may then denote $G \subseteq \prec$.

We will assume that the model $p$ is modular over $\prec$, $G$, $\theta$, and $\mathbf{x}$, or *order-modular* for short, in the sense of Koivisto and Sood (2004). The first part of the definition states that if $G$ is consistent with $\prec$, then

$$p(\prec, G) = \prod_{i=1}^{n} q_i(U_i) \rho_i(G_i)$$

where each $q_i$ and $\rho_i$ is some function from the subsets of $V - \{i\}$ to the nonnegative reals.[3] It should be noted that this prior does not respect Markov equivalence; for discussion, see Friedman and Koller (2003) and Koivisto and Sood (2004). The second part is a standard assumption (Cooper and Herskovits, 1992, Heckerman et al., 1995): given the structure $G$, the parameters $\theta$ decompose into independent local components $\theta_1, \ldots, \theta_n$, each $\theta_i$ depending only on the local structure $G_i$; furthermore, $p(x_i|x_{G_i}, G, \theta) = p(x_i|x_{G_i}, G_i, \theta_i)$.

Koivisto and Sood (2004) show that under an order-modular model and complete data it is possible to compute the joint posterior probability of any subnetwork (i.e., set of edges) in $O(n2^n)$ time, allowing exact Bayesian learning of networks with up to about 25 attributes. (A silent assumption is that one can efficiently evaluate the local marginal likelihood; closed-form expressions exist, e.g., for the Multinomial and the linear Gaussian model with conjugate priors.) We next review the key ingredients of the algorithm by Koivisto and Sood (2004) and fix some notation.

A structural feature, e.g., an edge, is conveniently represented by a modular indicator function, $f(G) = \prod_{i=1}^{n} f_i(G_i)$, where each $f_i(G_i)$ is either 1 or 0. For example, to represent an edge $(u, v)$, we set $f_v(G_v) = 1$ if and only if $u \in G_v$, while $f_i(G_i)$ is set to the constant 1 for all $i \neq v$ and all $G_i$. Then $f(G) = 1$ if and only if $G$ contains the edge $(u, v)$. Our chief interest is in evaluating the joint probability of the data and the feature, $p(\mathbf{x}, f)$, where $f$ is read as the event "$f(G) = 1$."

---

[1] Thanks to the improved algorithm, we are able to analyze several hundred random data sets with up to 25 attributes. With the same computational resources we would have had to restrict our experiments to networks of at most 16 attributes if we had used the original algorithm of Koivisto and Sood (2004).

[2] The linear order can be viewed as a technical device to construct a modular prior over DAGs.

[3] No normalization constant is needed, as any such factor can be absorbed into the functions $q_i$ and $\rho_i$. However, it makes sense to carry out the computations up to a normalization constant, as the constant is irrelevant for the posterior summaries (Koivisto and Sood, 2004).

We assign each family $(i, G_i)$, consisting of a node and its parents, the score

$$\beta_i(G_i) := \rho_i(G_i) p(\mathbf{x}_i | \mathbf{x}_{G_i}, G_i) f_i(G_i),$$

which quantifies the local goodness of $G_i$ as the parents of $i$. Further, it turns out to be convenient to transform the score of the actual parents $G_i$ to a score of the set of candidate parents $U_i$, defined as

$$\alpha_i(U_i) := q_i(U_i) \sum_{G_i \subseteq U_i} \beta_i(G_i).$$

We will discuss related Möbius transformation variants in detail in Section 4.

We can now express the joint probability of the data and the feature by a neat sum–product formula that has a simple recursive evaluation scheme:

$$p(\mathbf{x}, f) = \sum_{\prec} \prod_{i=1}^{n} \alpha_i(U_i) = L(V), \quad (2)$$

where the function $L$ is defined over all subsets of $V$ by setting $L(\emptyset) := 1$ and, recursively,

$$L(S) := \sum_{i \in S} \alpha_i(S - \{i\}) L(S - \{i\}).$$

The letter "$L$" anticipates the interpretation of this recursive procedure as the *left* (or *forward*) computation, as opposed to the *right* (or *backward*) computation to be introduced in the next section.

Finally, we note that the posterior probability of a feature $f$ is obtained as $p(f|\mathbf{x}) = p(\mathbf{x}, f)/p(\mathbf{x})$, where the denominator $p(\mathbf{x})$ can be computed like $p(\mathbf{x}, f)$ but for the trivial feature $f(G) \equiv 1$. As a consequence, the "priors" $\rho_i$ and $q_i$ need to be specified only up to constant factors. (To avoid using the proportionality sign, we assume that $\rho_i$ and $q_i$ absorb these constants.)

## 3 A FORWARD–BACKWARD ALGORITHM

We now focus on the following problem: Given a complete data set and an order-modular model $p$, compute for every edge $e$ the posterior probability that the network structure contains $e$. We note that this problem can be solved in $O(n^3 2^n)$ total time by running the algorithm of Koivisto and Sood (2004) separately for each edge.

However, computations for different edges involve a large proportion of overlapping elements. For example, switching from an edge $(u, v)$ to another edge $(u', v)$ corresponds to changing just the function $f_v$, hence affecting the function $\beta_v$ and, thereby, the function $\alpha_v$ only. Intuitively, it should be possible to exploit the overlap and reduce the total time requirement. We next describe how the calculations can be arranged so that the total running time reduces to $O(n 2^n)$.

Consider the sum over orders, given in (2). Our key idea is to compute and store not only the forward contribution $L(S)$, for all $S \subseteq V$, but also the *backward* or *right* contribution $R(T)$, for all $T \subseteq V$, defined by

$$R(T) := \sum_{\prec' \in \mathcal{L}(T)} \prod_{i \in T} \alpha_i((V - T) \cup U_i^{\prec'}).$$

In words, $R(T)$ is the contribution of the nodes in $T$, given that the nodes in $T$ are the $|T|$ last elements in the unknown linear order $\prec$ on $V$. Like $L$, the function $R$ can be evaluated recursively via the equations

$$R(T) = \sum_{i \in T} \alpha_i(V - T - \{i\}) R(T - \{i\})$$

and $R(\emptyset) = 1$.

To compute the sum over orders, we combine the left and right contributions. For any fixed node $v \in V$ (the endpoint of an edge), we break the sum over orders $\prec = (U_1, \ldots, U_n)$ into two nested sums: in the outer sum, we sum over $U_v$; in the inner sum we, conditionally on $U_v$, sum over the remaining sets $U_i$ for $i \neq v$ (denoted $U_{-v}$ for short). We observe that the inner sum further factorizes into the product of the left and the right term, $L(U_v) R(V - \{v\} - U_v)$. That is,

$$\begin{aligned} p(\mathbf{x}, f) &= \sum_{\prec} \prod_{i=1}^{n} \alpha_i(U_i) \\ &= \sum_{U_v} \alpha_v(U_v) \sum_{U_{-v}} \prod_{i \in V - \{v\}} \alpha_i(U_i) \\ &= \sum_{U_v} \alpha_v(U_v) L(U_v) R(V - \{v\} - U_v), \end{aligned}$$

where $U_v$ runs through all subsets of $V - \{v\}$.

At this point, it is enlighting to note that the above forward–backward expression already offers an $n$-fold speedup over the naive algorithm. To see this, suppose that we have precomputed the left and right contributions with respect to the trivial feature $f \equiv 1$. Then, to evaluate the posterior probability of any edge $(u, v)$ we only need to (i) recompute the function $\alpha_v$, and (ii) sum over all subsets $U_v$ of $V - \{v\}$. Step (i) takes $O(2^n)$ time; see Koivisto and Sood (2004) and the next section. Step (ii) takes $O(2^n)$ time as well. Thus, the total running time for all edges is $O(n^2 2^n)$.

In order to achieve another $n$-fold speedup, we plug in the definition of $\alpha_v(U_v)$, obtaining

$$p(\mathbf{x}, f) = \sum_{U_v} \left[ q_v(U_v) \sum_{G_v \subseteq U_v} \beta_v(G_v) \right] \times L(U_v) R(V - \{v\} - U_v).$$

Finally, reversing the order of summation yields

$$p(\mathbf{x}, f) = \sum_{G_v \subseteq V - \{v\}} \beta_v(G_v) \gamma_v(G_v),$$

where for all $G_v \subseteq V - \{v\}$ we define

$$\gamma_v(G_v) := \sum_{G_v \subseteq S \subseteq V - \{i\}} q_v(S) L(S) R(V - \{v\} - S).$$

It is useful to note that if $k$ is a fixed maximum indegree, then $\beta_v(G_v)$ vanishes whenever $G_v$ contains more than $k$ elements. In this case, the function $\gamma_v$ needs to be computed only at sets $G_v$ with at most $k$ elements.

So we have arrived at the following algorithm for computing the marginal posterior probabilities for all possible edges. Let the functions $\beta_i$, $\alpha_i$, $\gamma_i$, $L$ and $R$ be defined with respect to the trivial feature $f \equiv 1$, and let us denote the maximum indegree by $k$.

1. For all nodes $i \in V$ and subsets $G_i \subseteq V - \{i\}$ with $|G_i| \leq k$: compute $\beta_i(G_i)$.

2. For all $i \in V$ and $U_i \subseteq V - \{i\}$: compute $\alpha_i(U_i)$.

3. For all $S \subseteq V$: compute $L(S)$.

4. For all $T \subseteq V$: compute $R(T)$.

5. For all $v \in V$:

   (a) For all $G_v \subseteq V - \{i\}$ with $|G_v| \leq k$: compute $\gamma_v(G_v)$.

   (b) For all $u \in V - \{v\}$: compute the probability of the data $\mathbf{x}$ and the edge $e = (u, v)$, by

   $$p(\mathbf{x}, e) = \sum_{u \in G_v \subseteq V - \{v\}:|G_v| \leq k} \beta_v(G_v) \gamma_v(G_v),$$

   and output the posterior probability $p(e|\mathbf{x}) = p(\mathbf{x}, e)/L(V)$.

We characterize the running time of this algorithm under the assumption that the maximum indegree $k$ is a constant. Step 1 takes $O(n^{k+1})$ time. Step 2 takes $O(2^n)$ time as already mentioned; see the next section. Steps 3 and 4 can be computed in $O(2^n)$ time using the above given recursions. For each node $v \in V$, step 5(a) can be computed in $O(2^n)$ time, as we will show in the next section. For each $v$, step 5(b) takes $O(n^k)$ time. Thus, the total time requirement is $O(n2^n)$.

The following theorem summarizes the main result of this paper.

**Theorem 1 (Main)** *Let $\mathbf{x}$ be a complete data set over $n$ attributes, and let $p$ be an order-modular model. Then the marginal posterior probabilities for all the $n(n-1)$ edges can be evaluated in $O(n2^n)$ total time.*

## 4 THE FAST TRUNCATED MÖBIUS TRANSFORM

We have postponed to this section the discussion of certain summation formulas, namely those that define the functions $\alpha_i$ and $\gamma_i$. These summations can be viewed as Möbius transformations on a subset lattice. Below we elaborate on some recent results on the so called truncated Möbius transforms (Koivisto and Sood, 2004). For clarity of presentation, we introduce a generic notation—the connections to the notation and terms used in the previous section are implicit.

Let $N = \{1, \ldots, n\}$. Let $s: 2^N \to \mathbb{R}$ be a mapping from the subsets of $N$ onto the real numbers. We say that a function $t: 2^N \to \mathbb{R}$ is the (downward) Möbius transform of $s$ if

$$t(T) = \sum_{T \subseteq S \subseteq N} s(S) \qquad \text{for all } T \subseteq N.$$

A variant (the upward Möbius transform) is defined by replacing the summation conditions by "$S \subseteq T$". While the straightforward way to compute the Möbius transform takes $O(3^n)$ time, the fast Möbius transform algorithm takes only $O(n2^n)$ time; see, e.g., Kennes and Smets (1991).

We consider a scenario where we need to evaluate the function $t$ only at the sets $T$ that contain at most $k$ elements, given the function $s$ and a number $k \leq n$. We call the corresponding transform the *k-truncated downward Möbius transform*. In the remainder of this section we show that $O(k2^n)$ time suffices for evaluating such a transformation. Our result is dual to a similar results by Koivisto and Sood (2004): The upward Möbius transform can be computed in $O(2^n)$ time when $s(S)$ vanishes at all sets $S$ that contain more than a constant number ($k$) of elements.

*The fast truncated downward Möbius transform algorithm (FTDMT)* works as follows. We encode every subset $S \subseteq N$ bijectively by a 0-1 vector $(S_1, \ldots, S_n)$ where $S_i = 1$ if $i \in S$ and $S_i = 0$ otherwise; denote $S_{i:j}$ for $(S_i, S_{i+1}, \ldots, S_j)$. Consider the following algorithm. First, set $s_0(S_{1:n}) := s(S)$ for all $S \subseteq N$. Then, iteratively for $i = 1, \ldots, n$ transform the function $s_{i-1}$ to another function $s_i$ as follows: for all $T_1, \ldots, T_i$ and $S_{i+1}, \ldots, S_n$ in $\{0, 1\}$ satisfying $T_1 + \cdots + T_i \leq k$, compute

$$s_i(T_{1:i}, S_{i+1:n}) := \sum_{S_i = T_i}^{1} s_{i-1}(T_{1:i-1}, S_{i:n}).$$

We obtain by induction that $s_n(T_{1:n}) = t(T)$.

The running time of the above algorithm is proportional to the sum $A_1 + \ldots + A_n$, where $A_i$ is the number of 0-1 vectors $(T_{1:i}, S_{i+1:n})$ satisfying $T_1 + \cdots + T_i \leq k$. For $i \leq k$ we have $A_i = 2^n$, and for $i > k$ we have $A_i = 2^{n-i} B(i, k)$, where $B(i, k) = \sum_{j=0}^{k} \binom{i}{j}$.

To obtain sufficiently tight bounds for the terms $B(i, k)$, we use a well-known Chernoff bound (see, e.g., Hoeffding, 1963):

**Theorem 2 (Chernoff bound)** *Let $X_1, \ldots, X_n$ be any independent Poisson trial with $\Pr\{X_i = 1\} = \mu_i$ for all $i = 1, \ldots, n$. Let $X = \sum_{i=1}^{n} X_i$, $\mu = \sum_{i=1}^{n} \mu_i$, and $d > 0$. Then*

$$\Pr\{X \leq \mu - d\} \leq \exp\left[-\frac{d^2}{2\mu}\right].$$

We substitute $\mu_i := 1/2$ and $d := n/2 - k$, which gives us the following bound.

**Corollary 3** *If $n > 2k$, then*

$$\sum_{j=0}^{k} \binom{n}{j} \leq 2^n \exp\left[-\frac{n}{4} + k - \frac{k^2}{n}\right].$$

We use the above estimate to bound each $A_i$ for $i = 4k + l$ with $l \geq 0$. We have

$$\begin{aligned} A_i &= 2^{n-i} B(i, k) \\ &\leq 2^n \exp\left[-\frac{i}{4} + k - \frac{k^2}{i}\right] \\ &= 2^n \exp\left[-\frac{1}{4}\left(i - 4k + \frac{4k^2}{i}\right)\right] \\ &< 2^n \exp(-l/4). \end{aligned}$$

Finally, we sum up the bounds of $A_i$ for $i = 1, \ldots, n$. On one hand, $A_1 + \cdots + A_{4k-1} \leq (4k-1)2^n$, and on the other hand, $A_{4k} + \cdots + A_n < 2^n(e^0 + e^{-1/4} + e^{-2/4} + \cdots) = 2^n/(1 - e^{-1/4}) < 5 \cdot 2^n$. Combining these bounds gives $A_1 + \ldots + A_n < 4(k+1)2^n$. Thus, the problem of evaluating $t(T)$ at every $T$ of size at most $k$ can be solved in $O(k2^n)$ time.

We observe that the above analysis also applies to the fast truncated Möbius transform algorithm of Koivisto and Sood (2004), which—using the terminology introduced in this section—computes the $k$-truncated upward Möbius transform. Thus we have the following result, which also proves the conjecture posed by Koivisto and Sood (2004) regarding the precise role of $k$ in the time complexity.

**Theorem 4 (Fast truncated Möbius transforms)** *The $k$-truncated upward and downward Möbius transforms on the subset lattice of $n$ elements can be computed in $O(k2^n)$ time.*

## 5 EXPERIMENTAL RESULTS ON EDGE DISCOVERY

This section summarizes an experiment concerning the statistical power of discovering edges in Bayesian networks. Briefly, for various random BNs we (i) simulated data sets of different sizes, (ii) then computed the posterior probability of each possible edge, and (iii) finally summarized the discrepancy between the true network structure and the set of inferred edges.

The algorithms described in the previous sections have been implemented in the C++ language into a program named REBEL (Rapid Exact Bayesian Edge Learning).[4]

### 5.1 EDGE LEARNING AND ROC CURVES

We consider a scenario where one tries to learn the existence or nonexistence of as many edges as possible. For simplicity, we restrict our consideration to undirected edges, that is, we do not care if the orientation is incorrect.[5] Once a decision has been made for each possible edge, the outcome can be summarized by the number of true positive ($TP$), false positive ($FP$), true negative ($TN$), and false negative ($FN$) edges. For any loss (or utility) function of these numbers, we could derive an optimal Bayesian decision. However, rather than fixing any such function, we follow Husmeier (2003) and use the ROC (receiver operating characteristic) curve to summarize the learning performance against different possible loss functions. To this end, we let $\tau$ be a threshold parameter taking values in $[0, 1]$. Each undirected edge $\{u, v\}$ is claimed to be present in the network if and only if the posterior probability $p((u, v)|\mathbf{x}) + p((v, u)|\mathbf{x})$ exceeds $\tau$. For each value of $\tau$ we plot the sensitivity $TP/(TP + FN)$ (proportion of recovered true edges) against the complementary specificity $FP/(TN + FP)$ (proportion of false edges). Note that the constructed edge set is of course not intended to represent a valid BN structure.

### 5.2 SYNTHETIC DATA

We generated synthetic data from BN models with $n = 5, 10, 15, 20, 25$ nodes, the maximum indegree $k = 2, 3, 4, 5$, and $r = 2, 4$ states per variable. For each combination $(n, k, r)$, we generated 10 independent BN models by the following procedure.

---

[4] The program REBEL will be made publicly available at http://www.cs.helsinki.fi/u/mkhkoivi/.

[5] The data-generating model is not particularly informative about the direction of the edges. While entire network estimates (not used here) could be checked for Markov equivalence with the true network, checks for individual directed edges are somewhat problematic. Note that when the inferred undirected edges are compatible with the true network, we may expect (up to rare exceptions) that the correct Markov equivalence class is identified by the data.

1. Draw a linear order $\prec$ on the node set $\{1, \ldots, n\}$ uniformly at random (u.a.r.).

2. For each node $i$ (independently):
   (a) draw the number of parents of $i$, denoted as $n_i$, from $\{0, 1, \ldots, k\}$ u.a.r.;
   (b) draw the $n_i$ parents of $i$ from the predecessors of $i$, i.e., from $\{j : j \prec i\}$ u.a.r.;
   (c) for each value configuration of the parents (independently): draw a distribution on the states $\{1, \ldots, r\}$ from the uniform distribution (Dirichlet with all parameters set to 1).

Note that this procedure specifies an order-modular model $p_{n,k,r}$ for each configuration $(n, k, r)$.

From each of the resulting $5 \times 4 \times 2 \times 10 = 400$ BNs we generated 10,000 independent data points. Of these data sets we used nested subsets containing the first $m = 20, 100, 500, 2000,$ and $10,000$ data points.

### 5.3 MODELS FOR DATA ANALYSIS—ON A BAYESIAN NOTION OF POWER

Each data set drawn from model $p_{n,k,r}$ was also analyzed under $p_{n,k,r}$. This choice is motivated not only by simplicity, but also by a Bayesian interpretation of statistical power as an expectation over the prior. For example, the Bayesian power of learning an edge is the (prior) probability that the posterior guess about the presence of the edge will be correct.[6] Such probabilities w.r.t. a model $p_{n,k,r}$ can be estimated by Monte Carlo averages over a sample of realizations of network structures and data sets drawn from the model $p_{n,k,r}$. However, since we use ROC curves to summarize the learning ability, we do not explicitly estimate the power of learning any individual edge. That said, we will examine the distribution of ROC curves under each $p_{n,k,r}$, rather than computing any averages.

### 5.4 RESULTS

Figure 1 shows that networks of $n = 20$ attributes, with fixed maximum indegree $k$ and the number of states $r$, the power of edge discovery grows relatively smoothly with the number of data points, as expected. More interestingly, we also notice that increasing the maximum indegree has only a mild effect, whereas increasing the number of states seems to have a somewhat surprising effect: when there are 500 or more data points, edges can be more reliably discovered on 4-state attributes than on binary attributes. This can be explained by the smaller information content of binary attributes. The results for other number of attributes and other data samples (not shown) are qualitatively very similar.

---
[6]For a deeper discussion of the notion of Bayesian power and its connection to the notion of expected information gain, see Koivisto (2004, Chap. 2).

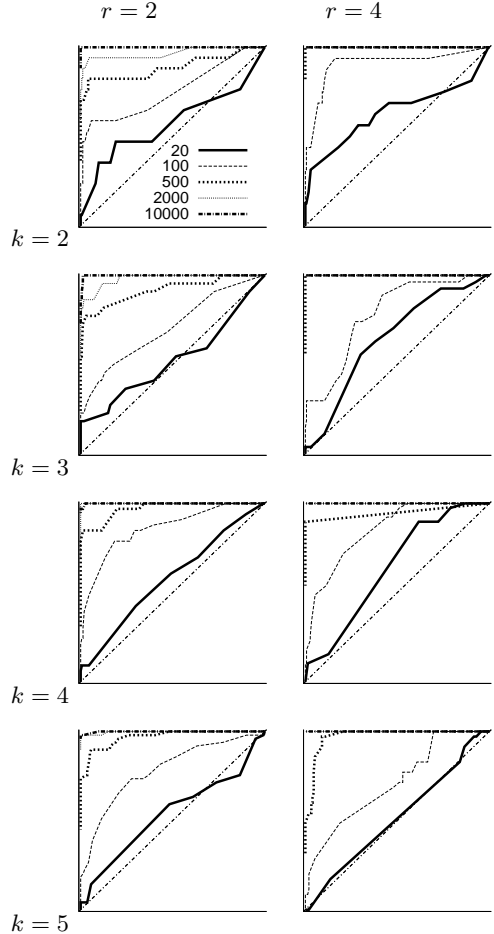

Figure 1: Power of edge discovery in networks of 20 attributes. For different maximum indegree $k$ and number of states $r$, the sensitivity ($y$-axis) is plotted against the complementary specificity ($x$-axis); the ranges of the axes are both $[0, 1]$. Each plot shows the ROC curve for five nested subsets of a single random data set. The straight diagonal line is the expected ROC curve of a random predictor.

As suggested by the results shown in Figure 2, there seems to be no clear increase nor decrease in power when the number of attributes grows. It should be noted, however, that Figure 2 (like Figure 1) shows results for one out of the 10 data samples per configuration. Yet, inspecting the results for all the 10 samples (not shown) does not change this view.

While Figures 1 and 2 display some "typical" results on the power of edge discovery, we get a more complete picture when we also explore the variation within the 10 ROC curves for each configuration $(n, k, r)$ and number of data points $m$ (Figures 3 and 4). We observe that the variance decreases as the number of data points increases. We also notice that, as already suggested, increasing the number of states $r$ from 2 to 4 decreases the average power for 100 or fewer data points, but slightly increases it for larger data sets.

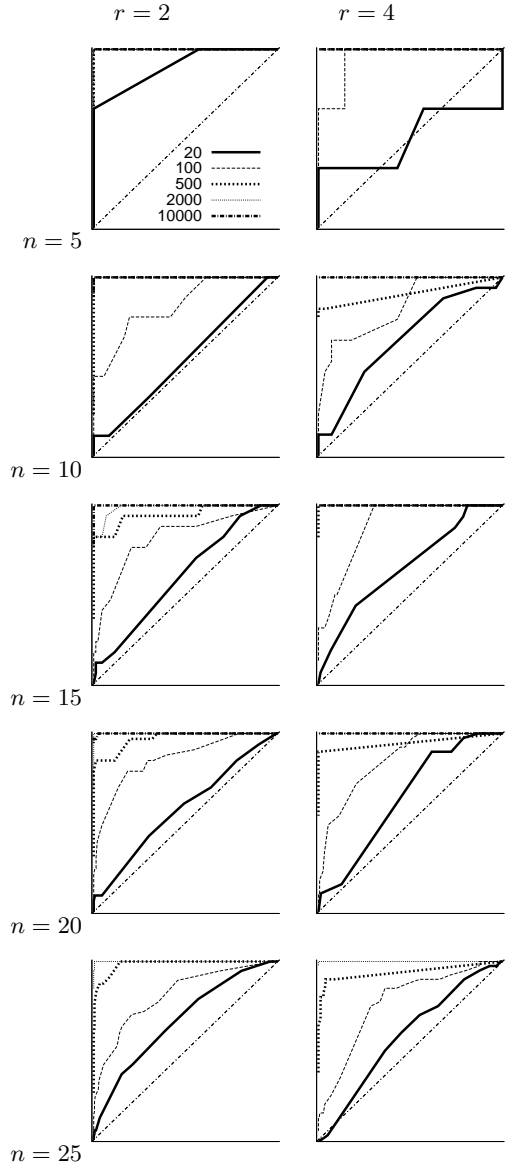

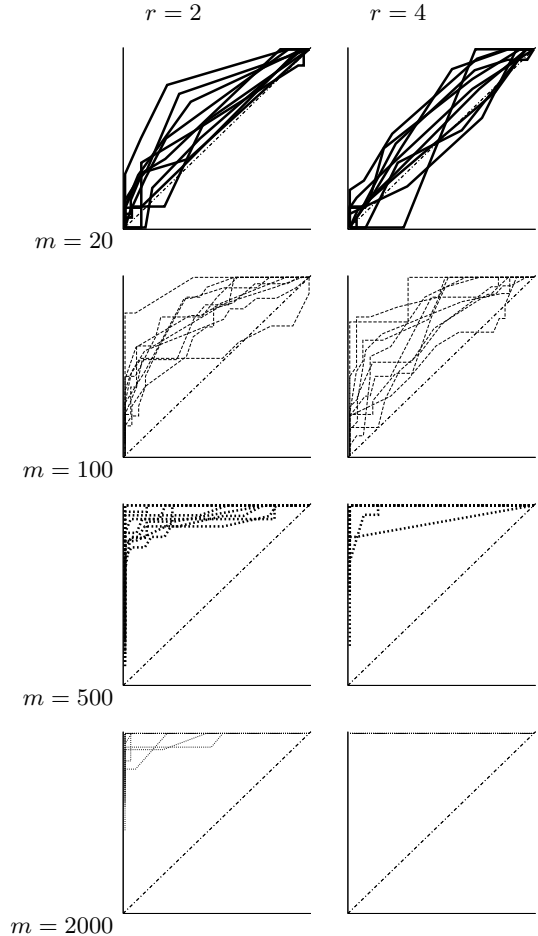

Figure 2: Power of edge discovery in networks with the maximum indegree 4. For different number of nodes $n$ and number of states $r$, the sensitivity ($y$-axis) is plotted against the complementary specificity ($x$-axis); the ranges of the axes are both $[0, 1]$. Each plot shows the ROC curve for five nested subsets of a single random data set. The straight diagonal line is the expected ROC curve of a random predictor.

Figure 3: Variance in the power of edge discovery in networks of 10 attributes with the maximum indegree 4. For different number of data points $m$ and number of states $r$, the sensitivity ($y$-axis) is plotted against the complementary specificity ($x$-axis); the ranges of the axes are both $[0, 1]$. Each plot shows the ROC curve for 10 random data set. The straight diagonal line is the expected ROC curve of a random predictor.

## 6  CONCLUDING REMARKS

We have presented a new exact algorithm for computing the marginal posterior probability of every edge in a Bayesian network. The algorithm runs in $O(n2^n)$ total time, where $n$ is the number of attributes, offering a 100- to 1000-fold speedup on the recent algorithm by Koivisto and Sood (2004).

We used this algorithm to study the statistical power of edge discovery. The advantage of the exact algorithm over inexact methods, e.g., MCMC, is that we do not have to speculate on the possible unreliability of the algorithm; instead, we can examine the pure combination of the model and the data.

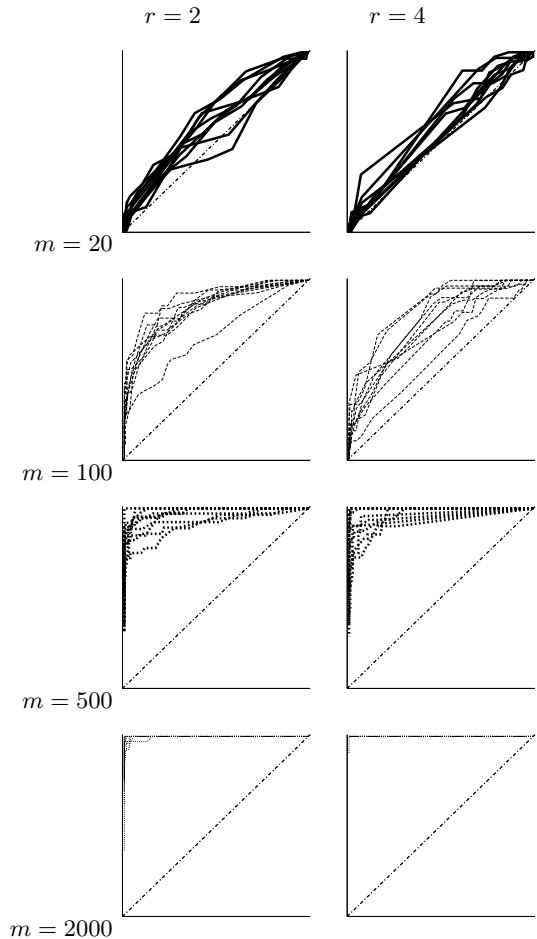

Figure 4: Variance in the power of edge discovery in networks of 20 attributes with the maximum indegree 4. For different number of data points $m$ and number of states $r$, the sensitivity ($y$-axis) is plotted against the complementary specificity ($x$-axis); the ranges of the axes are both $[0, 1]$. Each plot shows the ROC curve for 10 random data set. The straight diagonal line is the expected ROC curve of a random predictor.

There are also other potential applications. Prior sensitivity analysis concerns the robustness of learning results to perturbations of the model (prior). Inexact methods can be problematic, for one usually cannot tell apart the variances due to the model and due to the algorithm. Another application is the validation of heuristic methods by comparing the results produced by a heuristic algorithm to the exact results. Tests on, say, 25 attributes may reveal shortcomings of a heuristic and, vice versa, observing that a heuristic algorithm performs well on networks of this size suggests that it may do so on larger networks as well.

From the algorithmic point of view, the most important open problem is perhaps the issue of space complexity. Is it possible to reduce the $O(n2^n)$ space requirement of the presented algorithm, without sacrificing much in the running time?

## Acknowledgments

I wish to thank Evimaria Terzi and Kismat Sood for useful discussions.